\documentclass[lettersize,journal]{IEEEtran}
\usepackage{amsmath,amsfonts,amsthm}
\newtheorem{theorem}{Theorem}
\newtheorem{proposition}{Proposition}
\newtheorem{definition}{Definition}
\usepackage{algorithmic}
\usepackage{algorithm}
\usepackage{array}
\usepackage{subfigure}
\usepackage{textcomp}
\usepackage{stfloats}
\usepackage{url}
\usepackage{verbatim}
\usepackage{graphicx}
\usepackage{cite}
\usepackage{booktabs}
\begin{document}

\title{Differential equation and probability inspired graph neural networks for latent variable learning}

\author{Zhuangwei Shi${}^*$
\thanks{Zhuangwei Shi is with the College of Artificial Intelligence, Nankai University, Tianjin 300350, China.}
\thanks{${}^*$Correspondence: zwshi@mail.nankai.edu.cn}}

\markboth{Journal of \LaTeX\ Class Files,~Vol.~14, No.~8, August~2021}%
{Shell \MakeLowercase{\textit{et al.}}: A Sample Article Using IEEEtran.cls for IEEE Journals}


\maketitle

\begin{abstract}
    Probabilistic theory and differential equation are powerful tools for the interpretability and guidance of the design of machine learning models, especially for illuminating the mathematical motivation of learning latent variable from observation. Subspace learning maps high-dimensional features on low-dimensional subspace to capture efficient representation. Graphs are widely applied for modeling latent variable learning problems, and graph neural networks implement deep learning architectures on graphs. Inspired by probabilistic theory and differential equations, this paper conducts notes and proposals about graph neural networks to solve subspace learning problems by conditional random field and differential equation.
\end{abstract}


\section{Introduction}
\label{sec:introduction}

\IEEEPARstart{L}{atent} variable learning consists of algorithms that predict latent variable from observation. The subspace learning problem is defined from the motivation that predict latent variable from observation. The observed features $X$ are usually high-dimensional, yet machine learning models perform much better on analyzing low-dimensional representations. The latent representations span the low-dimensional subspace. The subspace may follow some constraints, usually low-rank constraints, sparse constraints and manifold constraints. Such constraints can be modeled through numerical optimization programming, but usually with probabilistic theory and differential equation as well. Appendix summarizes some classical subspace learning methods.

Subspace learning usually conducts manifold constraint. Manifold constraint is motivated by the \emph{manifold} in differential geometry, which can derive some remarkable partial differential equations (PDE). The Laplacian operator in PDE is correspond to the Laplacian matrix of a matrices. This idea can derive a series of ordinary differential equations (ODE) to describe the Laplacian-participated linear dynamic systems. Label propagation \cite{Zhou2004Learning,WangLabel,Wang2009Linear} follows manifold regularization \cite{Belkin2006Manifold}. For semi-supervised iterative algorithms such as label propagation algorithm and pagerank, establishing the equivalent relationship between its iterative equation and deep learning network can improve the effect of label propagation algorithm, and design semi-supervised iterative algorithms in deep learning network.

Graph neural networks \cite{ScarselliThe,KipfW16} (GNN) is widely applied to model graph relationship for deep learning \cite{shi2021vgaelda,jin2021lpigac,jin2022nimgsa}. Graphs are widely applied for modeling latent variable learning problems, and graph neural networks implement deep learning architectures on graphs. 

Vanilla GNN focus on classification task upon nodes on a graph. The graph convolutional networks (GCN) \cite{KipfW16}, GraphSAGE \cite{HamiltonInductive}, MPNN (message passing network) \cite{gilmer2017neural}, graph attention networks (GAT) \cite{VeliGraph}, graph autoencoders (GAE) \cite{kipf2016variational}, variational graph autoencoders (VGAE) \cite{kipf2016variational}, attention-based graph neural networks (AGNN) \cite{thekumparampil2018attentionbased}, graph isomorphism network (GIN) \cite{ginconv} and many other models were proposed. Subspace learning and differential equation guided graph neural networks \cite{deeperinsights2018,sgc2019} are useful for detecting subspace with low-rank or sparse property, or with better manifold smoothness. Probabilistic theory inspired graph neural networks \cite{kipf2016variational,shi2021vgaelda}, such as varitional inference and conditional random field, are powerful for detecting latent variable dependency and learning efficient representation.

\section{Differential equation and probability inspired graph neural networks}

\subsection{ODE inspired graph neural networks}

Deep learning and differential equations (dynamic systems) have certain equivalence under certain conditions. For example, ResNet can be regarded as the form of $x_{n+1}-x_n=f(x_n)$, which can be regarded as discretized $\dot x=f(x)$. For the optimal control form and feedback form of the differential (iterative) equation, the process of network training can be regarded as the process of solving the optimal control problem. Optimize the solution via Pontryagin's maximum principle (PMP) in optimal control \cite{Li2017Maximum}, or design a new deep learning network e.g. NeuralODE\cite{ChenNeural}, can improve the interpretability of the deep learning network.

When the graph neural network was first proposed, \cite{ScarselliThe} was given in the form of an iterative equation, denoting the feature as $X$, the hidden state as $H$, and the output as $O$, then $H\leftarrow F(H ,X),O\leftarrow G(H^*,X)$. For the current general graph neural network, its message passing process can also be written in the form of an iterative equation. For example, the GCN in \cite{KipfW16} can be written as (transductive learning)
\begin{equation}
    \bar{\mathbf{h}}_i^{(k)} \leftarrow \frac{1}{d_i + 1} \mathbf{h}_i^{(k-1)}+\sum_{j=1}^n \frac{1}{\sqrt{(d_i + 1) (d_j + 1)}}\mathbf{h}_j^{(k-1)},
\end{equation}
GraphSAGE \cite{HamiltonInductive} can be written as (inductive learning)
\begin{equation}
    \bar{\mathbf{h}}_i^{(k)} \leftarrow \frac{1}{d_i + 1} \left(\mathbf{h}_i^{(k-1)}+\sum_{j=1}^n \mathbf{h}_j^{(k-1)}\right), 
\end{equation}
PageRank \cite{pagerank} can be written as
\begin{equation}
    h_u = \frac{1-\alpha}{N} + \alpha \sum_{v \in \mathcal{N}(u)}\frac{h_v}{d_v}
\end{equation}
Among them, $\alpha$ is called the damping factor, which is generally taken as 0.85, $N$ is the number of nodes in the entire graph, and $d$ represents the degree of the node, which is similar to the message passing mechanism of GCN.

Consider the one-dimensional Weisfeiler-Lehman (WL-1) algorithm
\begin{equation}
    h^{(t+1)}_i \leftarrow \mathrm{hash}\left(\sum_{j\in\mathcal{N}_i} h^{(t)}_j\right) 
\end{equation}
Here $h_i^{(t)}$ represents the "color" (i.e. label information) of the node $v_i$ at the $t$th iteration, and $\mathcal{N}_i$ represents its neighborhood. Replace the hash operation with a neural network, then
\begin{equation}
    h^{(l+1)}_i = \sigma \left( \sum_{j\in\mathcal{N}_i} \frac{1}{c_{ij}}h^{(l)}_jW^{(l)} \right) \, 
\end{equation}

The discrete iteration is linked to dynamic programming. \cite{XuWhat} considered graph neural networks with the relaxation step of Bellman-Ford algorithm.
\begin{equation}
    d[t+1][i]=\min_{j\in\mathcal{N}_i} d[t][j] + w[j][i]
\end{equation}

Inspired by the Neural ODE\cite{ChenNeural}, \cite{cgnn2020} proposes a continuous graph neural network (continuous GNN), and uses the ordinary differential equation model to explain the GNN.

\subsection{PDE inspired graph neural networks}

\subsubsection{Heat conduction equation inspiration}

Use $\Delta$ to represent the Laplacian operator (harmonic operator), such as in the two-dimensional case
\begin{equation}
    \Delta=\frac{\partial^2}{\partial x^2}+\frac{\partial^2}{\partial y^2}
\end{equation}
Use $\Delta^2$ for the biharmonic operator, e.g. in the two-dimensional case
\begin{equation}
    \Delta^2=\left(\frac{\partial^2}{\partial x^2}+\frac{\partial^2}{\partial y^2}\right)^2=\frac{\partial^4}{\partial x^4}+2\frac{\partial^4}{\partial x^2\partial y^2}+\frac{\partial^4}{\partial y^4}
\end{equation}
On the graph $\Delta$ represents the normalized Laplacian matrix (the Laplacian operator on the discretized manifold), and the biharmonic operator $\Delta^2\approx \Delta^T \Delta$

\begin{theorem}\label{thm:yTLy}
    The following equality holds that
    \begin{equation}
        \min y^T \Delta y\Leftrightarrow\min\|\nabla y\|^2
    \end{equation}
\end{theorem}

Proof of all theorems are provided in Appendix.

\begin{definition}[Heat conduction equation]
    Heat $u$ on manifold $M$
    \begin{equation}
        \frac{\partial u}{\partial t}=\Delta u\label{eq:heat}
    \end{equation}
\end{definition}

\begin{theorem}\label{thm:rbf}
    Regard the label information as the temperature on the graph. When the "thermal equilibrium" is reached, that is, when the label propagation converges, there will be $\Delta u=0$. Based on this, the edge weights of graph can be set as RBF (radius basis function) kernel \cite{Belkin2001Laplacian,Ng2001On,Zhou2004Learning}.
    \begin{equation}
        w_{ij}=\exp\left(-\frac{\|x_i-x_j\|^2}{4t}\right)\label{eq:rbf}
    \end{equation}
\end{theorem}

Laplacian eigenmap \cite{Belkin2001Laplacian} defines edge weights according to \eqref{eq:rbf} to solve optimization problems
\begin{equation}
    \begin{split}
        \min & \,\, \mathrm{tr}(y^TLy)\\
        \mathrm{s.t.} & \,\, y^TDy=1,y^TDe=0
    \end{split}
\end{equation}
where $D, L$ represent the degree matrix and Laplacian matrix without regularization, and $e$ is an all-one vector. Let $0=\lambda_1<\lambda_2<...<\lambda_n$ be the eigenvalue of the generalized eigenvalue problem $Ly=\lambda Dy$, and take the $2\sim m+1$th element of each eigenvector to form $m$-dimensional features for each node.

Spectral clustering \cite{Ng2001On} is similar to \cite{Belkin2001Laplacian}, let $0=\lambda_1<\lambda_2<...<\lambda_n$ be $D^{-1/2}LD^{-1/2}$, take the $2\sim m+1$ element of each feature vector to form the $m$ dimension feature of each node, and then perform k-means clustering. \cite{ncut} proves that solving such an eigenvalue decomposition problem is equivalent to finding the optimal solution of a graph cut problem (i.e., normalized cut) (the sum of weights between classes is the smallest, and the sum of weights within classes is the largest).
\begin{equation}
    \begin{split}
        \min & \,\, Ncut=\frac{y^T(D-W)y}{y^T Dy}\\
        \mathrm{s.t.} & \,\, y^TDe=0
    \end{split}
\end{equation}
This is applied to image segmentation. When the value of each element in the eigenvector corresponding to the second smallest eigenvalue is taken out, the image is thresholded according to this value (or k-means clustering is performed on each pixel). class) to get the result of image segmentation.

After \cite{Wang2009Linear} introduces hyperedge, the problem of solving the weight matrix can be written as the following optimization problem
\begin{equation}
    \begin{split}
        \min_w & \,\, \varepsilon = \sum _ { i } \| x _ { i } - \sum _ { j : x _ { j } \in N ( x _ { i } ) } w _ { i j } x _ { j } \| ^ { 2 }\\
        \mathrm{s.t.} & \sum _ { j : x _ { j } \in N ( x _ { i } ) } w _ { i j }=1,w _ { i j }\ge 0
    \end{split}
\end{equation}

\begin{proposition}
    First-order CRF is linked to $\min y^T\Delta y$, second order is linked to $y^T\Delta^2 y$. (See Theorem \ref{thm:1stcrf} and \ref{thm:2ndcrf})
\end{proposition}

Now generalize \eqref{eq:heat} to the graph. 
\begin{definition}[Heat conduction equation on graph]
    For the heat $u_i$ of node $i$ we have
    \begin{equation}
         \Delta u_i=\sum_{j\in N(i)}w_{ij}u_j-d_iu_i
    \end{equation}
    so
    \begin{equation}
         \Delta u=(W-D)u=-Lu
    \end{equation}
    so
    \begin{equation}
         \partial u/\partial t=-Lu\label{eq:diffusion}
    \end{equation}
    Heat distribution $K_t=\exp(-tL)$, where $K_t(i,j)$ represents the heat conducted from $i$ to $j$ at time $t$. Kernel $K_t$ is equivalent to setting edge weights as \eqref{eq:rbf}.
\end{definition}

\begin{proposition}
    \cite{ZGL2003} suggests that the semi-supervised learning problem on the graph is equivalent to solving the harmonic equation $\Delta y=0$, and the boundary condition is the label of the marked point, which is equivalent to minimizing $y^T \Delta y$. The biharmonic operator is equivalent to minimizing $\|\Delta y\|^2$.
\end{proposition}

\subsubsection{PDE inspired spectral GNN}

\begin{proposition}
Green function of \eqref{eq:diffusion}
\begin{equation}
     G=\int_0^{+\infty}K_tdt=L^{-1}
\end{equation}
After introducing Green function, $y_u$ can be calculated like this
\begin{equation}
     y_u=G_{uu}W_{ul}y_l
\end{equation}
which is
\begin{equation}
     f(j)=\sum_{i=1}^l \sum_{k\in N(j)} y_i w_{ik} G(k,j)
\end{equation}
This is how to write $f$ as a kernel classifier.
\end{proposition}

\begin{definition}[Heat kernel]
    The solution to the equation \eqref{eq:diffusion} is
    \begin{equation}
         K_tu(x,0)=K_tf(x)=\int_M \kappa_t (x,y)f(y)dy
    \end{equation}
    Where $\kappa_t$ is called the heat kernel, and heat kernel has the following properties:
    \begin{equation}
         \kappa_t(x,y)=\sum_{i=1}^{+\infty}e^{-\lambda_i t}u_i^T(x)u_i(y)\label{eq:heatker}
    \end{equation}
    where $i$ is the $i$-th eigenvector of the Laplacian matrix.
\end{definition}

In addition, the spectrum of $K_t=\exp(-tL)$ can also be studied, and if $\lambda_i$ represents the eigenvalue of $L$, this group of basis $e^{-\lambda_it}$ can be investigated. In the frequency domain, that is, a set of basis $e^{-j\omega t}$, which is the idea of \emph{Fourier transform}. In fact, the Fourier transform is derived from the solution of the heat conduction equation.

\begin{proposition}
    Note that the first-order Taylor expansion of $e^{-\lambda_i}$ is $1-\lambda_i$, so it can be considered that $e^{-\lambda_it}\approx (1-\lambda)^t$, while $ (1-\lambda)^k$ corresponds to the $k$ order filter in the graph convolution.
\end{proposition}

Message passing process in the graph neural network can also be regarded as a process of "heat" propagation, and it can also be regarded as a kind of label propagation. \eqref{eq:heatker} is very similar to the Fourier transform on a graph. And Fourier transform on the graph is closely related to the graph neural network.

\subsubsection{PDE inspired spatial GNN}

Above, we have proposed an understanding of the graph neural network method in the spectral domain. 

For the graph neural network method in the spatial domain, for example, AGNN\cite{thekumparampil2018attentionbased} calculates $e_{ij}=\beta\cos(z_i,z_j)$, Where $\beta\in(0,1)$ is a learnable parameter, and then do softmax to get the weight
\begin{equation}
     w_{ij}=\frac{\exp e_{ij}}{\sum_{k\in\mathcal{N}(i)}\exp e_{ik}}
\end{equation}
This is very similar to the RBF kernel because
\begin{equation}
     \begin{split}
         e_{ij}&=-\frac{\|z_i-z_j\|^2}{2\sigma^2}\\&=-\frac{2(1-\cos(z_i,z_j))}{2 \sigma^2}\\&\overset{\text{def}}{=}-\beta (1-\cos(z_i,z_j)),\\ &=\beta\cos(z_i,z_j)-\beta
     \end{split}
\end{equation}
This is using the attention mechanism to learn edge weights similar to RBF.

\subsection{Graph neural networks inspired by differential equation}

\subsubsection{Label propagation and manifold learning}

A graph is the discretized form of a manifold, thus f can also be regarded as the discretized form of f, with its values equivalent to the values of f at the nodes of the graph. In such a viewpoint, the matrix I-W can be regarded as the graph Laplacian of this pasted graph intuitively. L is the Laplacian-Beltrami operator defined on the data manifold, and f is the function defined on this manifold.

\begin{proposition}\label{prop:manifold}
    \begin{enumerate}
        \item Semi-supervised learning problem on the graph is equivalent to solving the harmonic equation $\Delta f=0$, and the boundary condition is the label of the marked point, which is equivalent to minimizing $f^ T \Delta f$. Minimizing $f^T \Delta f$ is equivalent to minimizing $\|\nabla f\|^2$.
        \item Use $L$ to represent the regularized Laplacian matrix, in the spectral clustering \cite{Ng2001On}, convert the solution equation $Lf=0$ (that is, find the eigenvector of $L$) to minimize $ f^TLf$, and solving such an eigenvalue decomposition problem is equivalent to finding an optimal solution to a graph cut problem (i.e., normalized cut).
        \item Relax the solution equation $Lf=0$ (that is, find the eigenvector of $L$) to minimize $\|Lf\|^2=f^TL^TLf$, that is, solve the equation $L^TLf=0 $, that is, the biharmonic equation $\Delta^2 f=0$.
    \end{enumerate}
\end{proposition}

Label propagation can be regarded as a Laplacian least square method\cite{Belkin2006Manifold}, and its optimization objective is
\begin{equation}
    \min_F F^TLF+\frac{\gamma}{2}\|F-Y\|_F^2
\end{equation}
Considering that the regularized Laplace matrix can be written as $L=I-W$, where $W$ is the normalized edge weight, the optimal solution is $F^*=(1-\alpha)(I-\alpha W)^{-1}Y$, here $\alpha=1/(1+\gamma)$. This can be obtained by iterating $F^{(k)}=\alpha WF^{(k-1)}+(1-\alpha)Y,0<(1-\alpha)\ll 1$. Let $\eta=2/\gamma$, 
\[\begin{split}
    &\min_F\, \eta\mathrm{tr}(F^TLF)+\|YF\|^2\\ \Rightarrow\, &\eta LF+YF=0\\ \Rightarrow\, &F=(I+\eta L)^{-1}Y
\end{split}\]
And when the spectral radius of $\eta L$ is not greater than 1, the first-order Taylor expansions of $(I+\eta L)^{-1}$ and $e^{-\eta L}$ are both $I-\ eta L$. Use $F=W(W+\eta LW)^{-1}Y$ to find $F$ \cite{Xia2010Semi}, where $\eta$ is a hyperparameter, then
\[\begin{split}
    F&=W(W+\eta LW)^{-1}Y\\ &=W(W^{-1}-W^{-1}(\eta L)(I+\eta L)^{-1} )Y\\ &=(I-\eta L(I+\eta L)^{-1})Y\\ &=(I+\eta L)^{-1}Y
\end{split}\]

In this paper, when using GCN for node classification, the iterative form of GCN is
\begin{equation}
    H_{k+1}=(I+\eta L)^{-1}H_k\Theta_k,\eta>1
\end{equation}
It is propagated according to the label The formula $F=(I+\eta L)^{-1}Y$ improves the practice of GCN. This paper sets $\eta=2$.

Remark of this Theorem is on Appendix.

\subsubsection{Edge weight computation inspired by differential equation}

The label propagation algorithm is similar in form to the graph neural network. The edge weight of the label propagation algorithm is determined according to the distance between the features. The goal is to make the weight between the similar feature points as large as possible. In the graph neural network, the weight of the edge is set according to the actual problem, or learn it with the attention mechanism. e.g. \cite{VeliGraph} setting
\begin{equation}
    \begin{split}
        z_i^{(l)}&=\Theta^{(l)}h_i^{(l)} \\
        e_{ij}^{(l)}&=\text{LeakyReLU}(\vec a^{(l)^T}(z_i^{(l)}||z_j^{(l)}))\\
        w_{ij}^{(l)}&=\frac{\exp(e_{ij}^{(l)})}{\sum_{k\in \mathcal{N}(i)}^{}\ exp(e_{ik}^{(l)})}\\
        h_i^{(l+1)}&=\sigma\left(\sum_{j\in \mathcal{N}(i)} {w^{(l)}_{ij} z^{(l)} _j }\right)
    \end{split}
\end{equation}
Where $\Theta,a$ is the parameter learned by the neural network. AGNN\cite{thekumparampil2018attentionbased} uses the process of calculating the above $e_{ij}$ to be replaced by cosine similarity, $e_{ij}=\beta\cos(z_i,z_j)$, where $\beta\in(0 ,1)$ is a learnable parameter. MPNN (message passing network) \cite{gilmer2017neural} uses a fully connected layer to map the feature vector of the source node to the edge weight, ie $w_{ij}=FFNN(z_j)$.

Calculate $w_{ij}$ using RBF
\begin{equation}
     \begin{split}
         e_{ij}&=-\frac{\|z_i-z_j\|^2}{2\sigma^2}\\&=-\frac{2(1-\cos(z_i,z_j))}{2 \sigma^2}\\&\overset{\text{def}}{=}-\beta (1-\cos(z_i,z_j)),\\ w_{ij}&=\frac{\exp e_{ ij}}{\sum_{k\in\mathcal{N}(i)}\exp e_{ik}}
     \end{split}
\end{equation}
where $\beta\in(0,1)$ is a learnable parameter.

\begin{figure}
    \centering
    \includegraphics[width=0.28\textwidth]{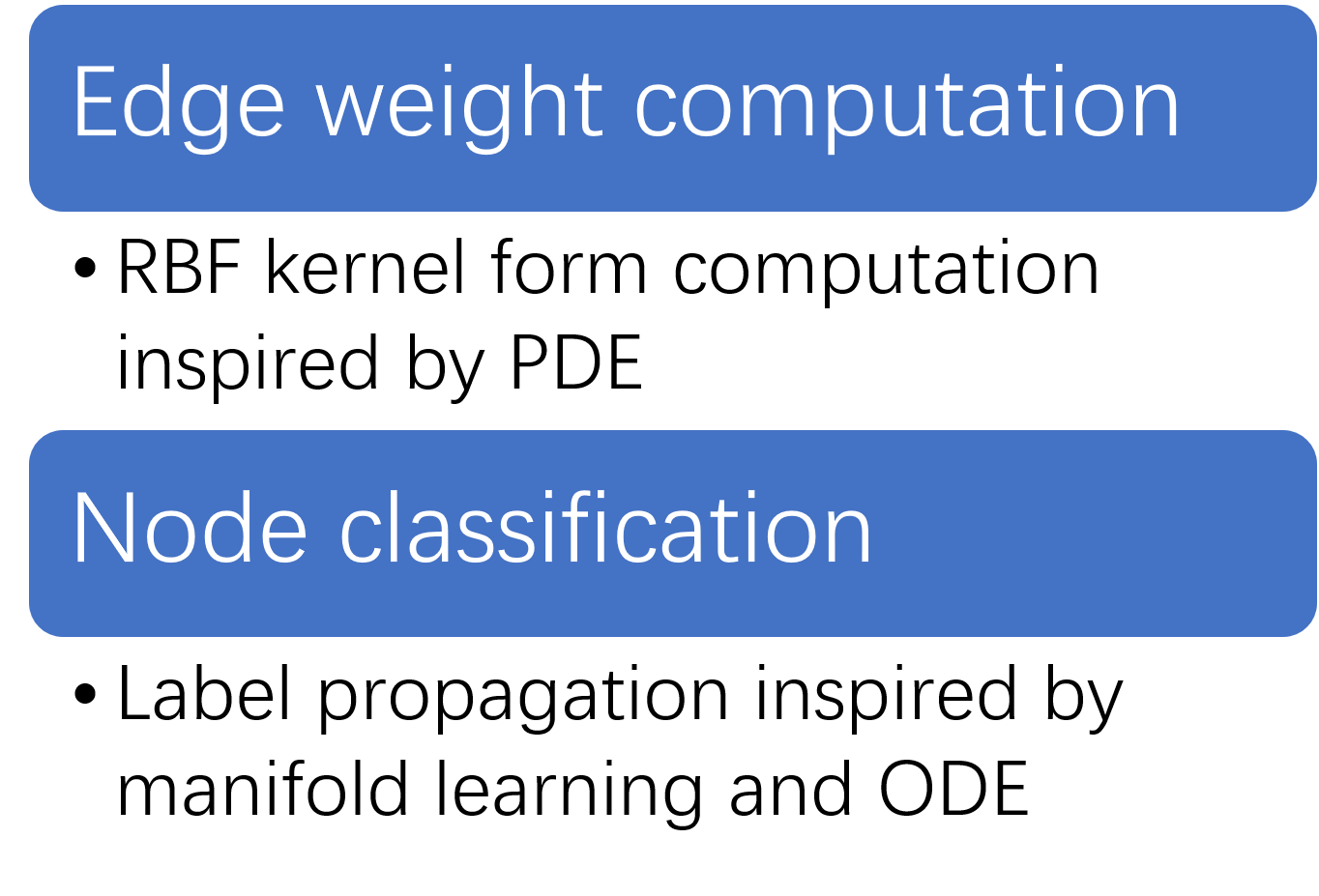}
    \caption{POGNN}\label{fig:pde-ode}
\end{figure}

The proposed PDE and ODE inspired graph neural networks is shown on Fig. \ref{fig:pde-ode}. It is so called POGNN. The iteration procedure $H_{k+1}=(I+\eta L)^{-1}H_k\Theta_k$ is inspired by ODE theory on label propagation, as well as manifold learning motivated by PDE. The RBF and neural network based edge weights computation is also inspired by PDE and ODE.

\subsection{Conditional random fieid for different equation inspired graph neural networks}

Consider each node $v_i$ on the graph as a factor $x_i$ in the factor graph, remember $x=(x_1,...,x_n)^T$, compatibility function
\begin{equation}
    \psi(x)=\exp (-E(x))
\end{equation}
Energy function
\begin{equation}
    E(x)=\frac{1}{2}(x-\mu)^TQ(x-\mu)
\end{equation}
Joint probability
\begin{equation}
    p(x)=\psi(x)/Z
\end{equation}
where normalized factor$Z=(2\pi)^{n/2}|Q|^{-1/2}$. Matrix $Q\in\mathbb{R}^{n\times n}$ is precision matrix, i.e. the inverse of covariance matrix. The normalized Laplacian matrix $L$ of a graph can be set as a precision matrix.

\begin{theorem}\label{thm:1stcrf}
    For the conditional random field, when the increments are first-order, the energy function is the quadratic form of harmonic operator (normalized Laplacian matrix).
    \begin{equation}
        E=\frac { 1 } { 2} x ^ { T } L x = \frac { 1 } { 2 }\sum_{i,j} w _ { i j } ( x _ { i } - x _ { j } ) ^ { 2 } 
    \end{equation}
\end{theorem}

\begin{theorem}\label{thm:2ndcrf}
    For the conditional random field, when the increments are second-order, the energy function is the quadratic form of biharmonic operator.
    \begin{equation}
        E=\frac { 1 } { 2} x ^ { T } L^TL x = - \frac { 1 } { 2 } \sum _ { i } ( \sum _ { j \in N _ { i } } (w _ { i j } x _ { j } - x _ { i } ) ^ { 2 } )
    \end{equation}
\end{theorem}

This finding explains proposition \ref{prop:manifold}, leading to compare $H_{k+1}=(I+\eta L)^{-1}H_k\Theta_k$ and $H_{k+1}=(I+\eta L^TL)^{-1}H_k\Theta_k$.

\section{Results}

Cora dataset\cite{2008Collective} has a total of 2708 sample points, each sample point is a paper, and all sample points are divided into 7 categories. Each paper is represented by a 1433-dimensional word vector to represent its feature vector. Each element of the word vector corresponds to a word, and the element has only two values of 0 or 1. Take 0 to indicate that the word corresponding to this element is not in the paper, and take 1 to indicate that it is in the paper. The citation relationship of the paper is the edge of the graph, and there are no isolated points on the graph. Only a small number of sample points (140 points) in the dataset are labeled with class labels, so the classification problem belongs to semi-supervised classification.

Conduct the following experiments on the Cora dataset, setting hidden dim=16, learning rate=0.1, weight decay = 5e-4, epoch=50. The propagation process $H_{k+1}=AH_{k}\Theta_k$, and results of using different $A$ are shown in Table \ref{tab:a}.

\begin{table}[]
    \centering
    \caption{Results using different $A$}
    \label{tab:a}
    \begin{tabular}{@{}cccc@{}}
    \toprule
    $\eta$ & $A=I-\eta L$ & $A=(I+\eta L)^{-1}$ & $A=(I+\eta L^TL)^{-1}$ \\ \midrule
    0.5 & 0.785 & 0.757 & 0.720\\ 
    1 & 0.827 & 0.812 & 0.776\\
    1.5 & 0.665 & 0.826 & 0.787\\
    2 & 0.449 & \textbf{0.837} & 0.795\\
    5 & 0.436 & 0.834 & 0.781\\ \bottomrule
    \end{tabular}
\end{table}

Therefore, $A=(I+\eta L)^{-1}$ performs best when $\eta=2$.

\begin{table}[]
    \centering
    \caption{Comparison with other models}
    \label{tab:tentative}
    \begin{tabular}{@{}ccccc@{}}
    \toprule
    Method & GCN & GAT & GraphSAGE & POGNN\\ \midrule
    Acc & 0.813 & 0.820 & 0.828 & 0.837\\ \bottomrule
    \end{tabular}
\end{table}

\section{Conclusions}

Probabilistic theory and differential equations are essential for understanding and guiding the design of machine learning models, particularly for understanding the mathematical basis of learning latent variables from observations. Subspace learning maps high-dimensional features onto low-dimensional subspaces to create an efficient representation. Graphs are often used to model latent variable learning problems, and graph neural networks implement deep learning architectures on graphs. This paper draws on probabilistic theory and differential equations to provide insights and suggestions on how graph neural networks can be used to solve subspace learning problems through conditional random field and differential equations.

\appendices

\section{Proof of theorems}

Proof of Theorem \ref{thm:yTLy}

\begin{proof}
    \begin{equation}
        \Delta(y)=\nabla\cdot\nabla y=\mathrm{div \nabla y},\quad \int_M<X,\nabla y>=\int_M \mathrm{div}(X)y
    \end{equation}
    So
    \begin{equation}
        \int_M \|\nabla y\|^2=\int_M \Delta(y)y
    \end{equation}
    So
    \begin{equation}
        \min y^T \Delta y\Leftrightarrow\min\|\nabla y\|^2
    \end{equation}
\end{proof}

Proof of Theorem \ref{thm:rbf}

\begin{proof}
    We regard the label information as the temperature on the graph. When the "thermal equilibrium" is reached, that is, when the label propagation converges, there will be $\Delta u=0$. Let $u(x,0)=f(x)$ be the initial heat distribution, and solve the problem of partial differential equation \eqref{eq:heat} according to this initial condition, which is called the Cauchy problem of heat conduction equation. The special solution of \eqref{eq:heat} when $f(x)=\delta(x)$ is
    \begin{equation}
        \Phi_t(x)=\frac{1}{(4\pi t)^{n/2}}\exp\left(-\frac{\|x\|^2}{4t}\right),t>0
    \end{equation}
    This is also known as the fundamental solution of the heat conduction equation, and
    \begin{equation}
        \int_{\mathbb{R}^n}\Phi_t(x)dx=1
    \end{equation}
    Let $H_t(x,y)=\Phi_t(x-y)$, the solution of \eqref{eq:heat} is
    \begin{equation}
        u(x,t)=\int_M H_t(x,y)f(y)dy
    \end{equation}
    That is, the convolution of the kernel $H_t$ and the initial condition $f(x)$. so
    \begin{equation}
        \begin{split}
            \Delta f(x)&=\frac{\partial}{\partial t}\left(\int_M H_t(x,y)f(y)dy\right)\bigg|_{t\to 0}\\
            &=-\frac{1}{t}\left(f(x)-g(x,y)\right)
        \end{split}
    \end{equation}
    \begin{equation}
        g(x,y)=(4\pi t)^{-n/2}\int_M \exp\left(-\frac{\|x-y\|^2}{4t}\right)f(y)dy
    \end{equation}
    For a point $x_i$, assuming that there are $k$ points in its neighborhood, then
    \begin{equation}
        \Delta f(x_i)=-\frac{1}{t}\left(f(x_i)-g(x_i,x_j)\right)
    \end{equation}
    \begin{equation}
        g(x_i,x_j)=\frac{1}{k}(4\pi t)^{-n/2}\sum_{j\in N(i)} \exp\left(-\frac{\|x_i-x_j\|^2}{4t}\right)f(x_j)
    \end{equation}
    where $N(i)=\{x_j|0<\|x_i-x_j\|<\varepsilon\}$, when $\Delta f(x_i)\approx 0$
    \begin{equation}
        \frac{1}{k}(4\pi t)^{-n/2}\sum_{j\in N(i)} \exp\left(-\frac{\|x_i-x_j\|^2}{4t}\right)=1
    \end{equation}
    so
    \begin{equation}
        w_{ij}=\exp\left(-\frac{\|x_i-x_j\|^2}{4t}\right)
    \end{equation}
\end{proof}

Proof of Theorem \ref{thm:1stcrf}

\begin{proof}
First-order increment
\begin{equation}
    \dot x_{ij}=x_i/\sqrt{\sum_i w_{ji}}-x_j/\sqrt{\sum_j w_{ij}}=x_i-x_j
\end{equation}
If $e$ is an all-one vector, then $Q$ satisfies $Qe=0$, and because $\sum_j w_{ij}=1$, it can be assumed that $d _ { ij } \sim N ( 0,1 / w _ { ij } )$, where $w _ { ij }$ represents the edge weights, assuming that the increments are independent, the joint probability distribution of $x$
\begin{equation}
    \begin{aligned} p ( x ) &{\propto \prod _ { ( i , j ) \in E } p ( \dot x_ { i j } ) }\\&{ \propto \prod _ { ( i , j ) \in E } \exp ( - \frac { 1 } { 2 } w _ { i j } ( x _ { i } - x _ { j } ) ^ { 2 } ) }\\&{ = \exp ( - \frac { 1 } { 2 } \sum _ { ( i , j ) \in E } w _ { i j } ( x _ { i } - x _ { j } ) ^ { 2 } ) }\\&{ = \exp ( - \frac { 1 } { 2
    } x ^ { T } Lx ) }\end{aligned}
\end{equation}
\end{proof}

Proof of Theorem \ref{thm:2ndcrf}

\begin{proof}
For a graph with $n$ nodes, we introduce $n$ hyperedge nodes, and the hyperedge node $e_i$ contains all nodes $v_i$ and $v_i$'s neighborhood set $N(i)$ node. The labels of the hypernodes are weighted and summed by the node labels in $N(i)$, that is, $\sum _ { j \in N _ { i } } w _ { ij } y _ { j } $, so for the second order , the increment is defined as the difference between the hypernode and its corresponding node
\begin{equation}
    \ddot x_ { i } = x _ { i } - \sum _ { j \in N _ { i } } w _ { i j } x _ { j }
\end{equation}
where $\sum _ { j \in N _ { i } } w _ { i j } = 1$. For the precision matrix $Q^h=\mathrm{diag}\{\ddot x_i\}$ of the hypergraph, there is still $Q^he=0$, and $e$ is an all-one vector, so it can be assumed that $d_i\sim N( 0,1)$, then
\begin{equation}
    \begin{aligned} p ( y )&{\propto \prod _ { i } p ( \ddot x_ { i } ) }\\&{ \propto \exp ( - \frac { 1 } { 2 } \sum _ { i } ( \sum _ { j \in N _ { i } } (w _ { i j } x _ { j } - x _ { i } ) ^ { 2 } ) }\\&{ = \exp ( - \frac { 1 } { 2 } x ^ { T } ( I - W ) ^ { T } ( I - W ) x ) }\\&=\exp(-\frac { 1 } { 2 }x^TL^TLx)\end{aligned}
\end{equation}
\end{proof}

\bibliographystyle{IEEETrans}
\bibliography{reference}

\begin{thebibliography}{10}
\providecommand{\url}[1]{#1}
\csname url@samestyle\endcsname
\providecommand{\newblock}{\relax}
\providecommand{\bibinfo}[2]{#2}
\providecommand{\BIBentrySTDinterwordspacing}{\spaceskip=0pt\relax}
\providecommand{\BIBentryALTinterwordstretchfactor}{4}
\providecommand{\BIBentryALTinterwordspacing}{\spaceskip=\fontdimen2\font plus
\BIBentryALTinterwordstretchfactor\fontdimen3\font minus
  \fontdimen4\font\relax}
\providecommand{\BIBforeignlanguage}[2]{{%
\expandafter\ifx\csname l@#1\endcsname\relax
\typeout{** WARNING: IEEEtranS.bst: No hyphenation pattern has been}%
\typeout{** loaded for the language `#1'. Using the pattern for}%
\typeout{** the default language instead.}%
\else
\language=\csname l@#1\endcsname
\fi
#2}}
\providecommand{\BIBdecl}{\relax}
\BIBdecl

\bibitem{Belkin2001Laplacian}
M.~Belkin and P.~Niyogi, ``Laplacian eigenmaps and spectral techniques for
  embedding and clustering,'' \emph{Advances in Neural Information Processing
  Systems}, vol.~14, no.~6, pp. 585--591, 2001.

\bibitem{Belkin2006Manifold}
M.~Belkin, P.~Niyogi, and V.~Sindhwani, ``Manifold regularization: A geometric
  framework for learning from labeled and unlabeled examples,'' \emph{Journal
  of Machine Learning Research}, vol.~7, no.~1, pp. 2399--2434, 2006.

\bibitem{ChenNeural}
R.~T.~Q. Chen, Y.~Rubanova, J.~Bettencourt, and D.~Duvenaud, ``Neural ordinary
  differential equations,'' in \emph{NeurIPS}, 2018.

\bibitem{gilmer2017neural}
J.~Gilmer, S.~S. Schoenholz, P.~F. Riley, O.~Vinyals, and G.~E. Dahl, ``Neural
  message passing for quantum chemistry,'' in \emph{ICML}, 2017.

\bibitem{HamiltonInductive}
W.~L. Hamilton, R.~Ying, and J.~Leskovec, ``Inductive representation learning
  on large graphs,'' in \emph{NeurIPS}, 2017.

\bibitem{jin2022nimgsa}
C.~Jin, Z.~Shi, K.~Lin, and H.~Zhang, ``Predicting mirna-disease association
  based on neural inductive matrix completion with graph autoencoders and
  self-attention mechanism,'' \emph{Biomolecules}, vol.~12, no.~1, p.~64, 2022.

\bibitem{jin2021lpigac}
C.~Jin, Z.~Shi, H.~Zhang, and Y.~Yin, ``Predicting lncrna–protein
  interactions based on graph autoencoders and collaborative training,'' in
  \emph{IEEE International Conference on Bioinformatics and Biomedicine (BIBM),
  Houston, USA, 9-12 December}, 2021.

\bibitem{KipfW16}
T.~Kipf and M.~Welling, ``Semi-supervised classification with graph
  convolutional networks,'' in \emph{ICLR}, 2017.

\bibitem{kipf2016variational}
T.~N. Kipf and M.~Welling, ``Variational graph auto-encoders,'' in
  \emph{NeurIPS}, 2016.

\bibitem{Li2017Maximum}
Q.~Li, C.~Long, T.~Cheng, and W.~E, ``Maximum principle based algorithms for
  deep learning,'' \emph{Journal of Machine Learning Research}, vol.~18, no.~1,
  2017.

\bibitem{deeperinsights2018}
Q.~Li, Z.~Han, and X.-M. Wu, ``Deeper insights into graph convolutional
  networks for semi-supervised learning,'' in \emph{AAAI}, 2018.

\bibitem{Ng2001On}
A.~Y. Ng, M.~I. Jordan, and Y.~Weiss, ``On spectral clustering: Analysis and an
  algorithm,'' in \emph{NeurIPS}, 2001.

\bibitem{pagerank}
L.~Page, S.~Brin, R.~Motwani, and T.~Winograd, ``The pagerank citation ranking:
  Bringing order to the web,'' Stanford University, Tech. Rep., 1999.

\bibitem{ScarselliThe}
F.~Scarselli, M.~Gori, A.~C. Tsoi, M.~Hagenbuchner, and G.~Monfardini, ``The
  graph neural network model,'' \emph{IEEE Transactions on Neural Networks},
  vol.~20, no.~1, pp. 61--80, 2009.

\bibitem{2008Collective}
P.~Sen, G.~Namata, M.~Bilgic, L.~Getoor, B.~Galligher, and T.~E. Rad,
  ``Collective classification of network data,'' \emph{AI Magazine}, vol.~29,
  no.~3, p.~93, 2008.

\bibitem{ncut}
J.~Shi and J.~Malik, ``Normalized cuts and image segmentation,'' \emph{IEEE
  Transactions on Pattern Analysis and Machine Intelligence}, vol.~22, no.~8,
  pp. 888--905, 2000.

\bibitem{shi2021vgaelda}
Z.~Shi, H.~Zhang, C.~Jin, X.~Quan, and Y.~Yin, ``A representation learning
  model based on variational inference and graph autoencoder for predicting
  lncrna-disease associations,'' \emph{BMC Bioinformatics}, vol.~22, p. 136,
  2021.

\bibitem{thekumparampil2018attentionbased}
K.~K. Thekumparampil, C.~Wang, S.~Oh, and L.-J. Li, ``Attention-based graph
  neural network for semi-supervised learning,'' \emph{arXiv preprint
  arXiv:1803.03735}, 2018.

\bibitem{VeliGraph}
P.~Velickovic, G.~Cucurull, A.~Casanova, A.~Romero, P.~Lio, and Y.~Bengio,
  ``Graph attention networks,'' in \emph{ICLR}, 2018.

\bibitem{WangLabel}
F.~Wang and C.~Zhang, ``Label propagation through linear neighborhoods,'' in
  \emph{ICML}, 2006.

\bibitem{Wang2009Linear}
J.~Wang, F.~Wang, C.~Zhang, H.~C. Shen, and L.~Quan, ``Linear neighborhood
  propagation and its applications,'' \emph{IEEE Transactions on Pattern
  Analysis and Machine Intelligence}, vol.~31, no.~9, pp. 1600--1615, 2009.

\bibitem{sgc2019}
F.~Wu, T.~Zhang, A.~H. de~Souza, C.~Fifty, T.~Yu, and K.~Q. Weinberger,
  ``Simplifying graph convolutional networks,'' in \emph{ICML}, 2019.

\bibitem{cgnn2020}
L.-P. A.~C. Xhonneux, M.~Qu, and J.~Tang, ``Continuous graph neural networks,''
  in \emph{ICML}, 2020.

\bibitem{Xia2010Semi}
Z.~Xia, L.~Y. Wu, X.~Zhou, and S.~T.~C. Wong, ``Semi-supervised drug-protein
  interaction prediction from heterogeneous biological spaces,'' \emph{Bmc
  Systems Biology}, vol.~4, no. Suppl 2, p.~S6, 2010.

\bibitem{ginconv}
K.~Xu, W.~Hu, J.~Leskovec, and S.~Jegelka, ``How powerful are graph neural
  networks?'' in \emph{International Conference on Learning Representations},
  2019.

\bibitem{XuWhat}
K.~Xu, J.~Li, M.~Zhang, S.~S. Du, K.-i. Kawarabayashi, and S.~Jegelka, ``What
  can neural networks reason about?'' in \emph{ICLR}, 2020.

\bibitem{Zhou2004Learning}
D.~Zhou, O.~Bousquet, T.~N. Lal, J.~Weston, and B.~Scholkopf, ``Learning with
  local and global consistency,'' \emph{Advances in neural information
  processing systems}, vol.~16, no.~3, 2004.

\bibitem{ZGL2003}
X.~Zhu, J.~Lafferty, and Z.~Ghahramani, ``Semi-supervised learning using
  gaussian fields and harmonic functions,'' in \emph{ICML}, 2003, pp. 912--919.

\end{thebibliography}

\vfill

\end{document}